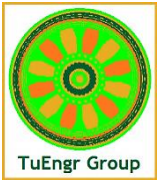

**International Transaction Journal of Engineering, Management, & Applied Sciences & Technologies**

**http://TuEngr.com**

TuEngr Group

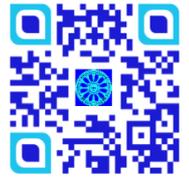

# AN EXPERIMENT ON MEASUREMENT OF PAVEMENT ROUGHNESS VIA ANDROID-BASED SMARTPHONES

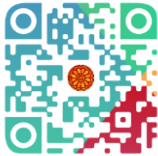

**Piyasak Thiandee [a], Boonsap Witchayangkoon [a*], Sayan Sirimontree [a], Ponlathep Lertworawanich [b]**

[a] *Department of Civil Engineering, Thammasat School of Engineering, Thammasat University, THAILAND*
[b] *Bureau of Highways Maintenance Management, Department of Highways, Ministry of Transportation, Royal Thai Government, THAILAND*

| A R T I C L E I N F O | A B S T R A C T |
|---|---|
|  | The study focuses on the experiment of using three different smartphones to collect acceleration data from vibration for the road roughness detection. The Android operating system is used in the application. The study takes place on asphaltic pavement of the expressway system of Thailand, with 9 km distance. The run vehicle has an inertial profiler with laser line sensors to record road roughness according to the IRI. The RMS and Machine Learning (ML) methods are used in this study. There is different ability of each smartphone to detect the vibration, thus different detected values are obtained. The results are compared to the IRI observation. ML method gives better result compared to RMS. This study finds little relationship between IRI and acceleration data from vibration collected from smartphones.<br>© 2019 INT TRANS J ENG MANAG SCI TECH. |

## 1. INTRODUCTION

Smoothness of road pavement is an important driving factor for vehicle as it reflects comfort-driving experience to vehicle driver and passengers. The term smoothness, on the other hand, indicates the roughness of the road surface. The vehicle-pavement system has been defied by ASTM E867 (2012), linking to the satisfactory of road users. This also indicates various road factors such as road performance, related accidents, vehicular maintenances, and fuel consumptions.

Pavement smoothness is controlled from road construction phase. The follow-up multiple pavement surveys are done to continuously monitor and evaluate the quality of the road pavement surface. Many available tools make it possible to find and measure road smoothness (FHWA, 2016). Straightedge is a simple tool to check the road flatness at a particular point. Inertial profilers with laser line sensors can be used for high-speed pavement profiling, giving ability to quickly obtained the surface roughness for hundred points along the laser line. Pavement 3D profiles can be generated with reliability (Pastorius et al., 2012) and roughness can be detected.

## 2. LITERATURE REVIEW

There are many studies on road roughness detection regarding measuring systems and calibrations (e.g. Gillespie, 1980; Cantisani, 2010). However, normally these studied used expensive devices. With the advent of smartphone with low cost and being used by everyone and the ease of app writing technology, it is possible to apply the smartphone and related technology to possible detect and quantify road roughness. There are many studies that try to apply smartphone to detect road roughness.

Hanson et al. (2014) used accelerometer data from mobile phone to detect road roughness by installing the phones (Samsung Galaxy SIII and iPhone 5) at the windshield of compact and SUV cars. With multi-processing to convert the measured data to IRI values and vehicular speeds (50 and 80 km/h). This study found it was possible to use smartphones for road roughness detections.

Aleadelat et al. (2018) applied the vertical acceleration data from smartphones (a Sony Xperia A and a Samsung Galaxy S III) to correlate with IRI via signal processing and pattern recognition techniques. The initial results give good levels of reasonable certainty.

Bisconsini et al. (2019) used smartphones to evaluate pavement roughness due to their cheap cost. A smartphone was placed on the car dashboard, at different running speeds. When compared with to the IRI, RMS for vertical acceleration gave a high correlation with IRI.

Yeganeh et al. (2019) used smartphones to observe asphaltic pavement urban road networks. Pavement Distress Index (PDI) has been considered. ANOVA (Analysis of variance) was used. The result found medium correlations between smartphone roughness measurement and IRI values.

With the advent of mobile phone and related technology, there have been significant advances in technology and application software. With the running of application software, signal detector is likely possible with smartphone and smartwatch. This research tries to take advantages of smartphone, by doing measuring accelerations from vibration detection caused by road roughness while driving a vehicle and assessment of the relationships between International Roughness Index (IRI) and such data. This study spotlights on the field experiment with different smartphones that have different CPU and accelerometer.

## 3. METHOD

### 3.1 MOBILE APPLICATION

The application software used in this research is a newly developed application, called Road Surface Survey (RSS) App (Figure 2). RSS App works on Android®, the world's most popular mobile platform. The main function is to detect acceleration from vibration by using frequency at 100 Hz. The smartphone accelerometer model, Figure 1, senses the movement from each axis. The captured data in the central processing unit is recorded via the RSS App. This study experiments on three smartphones from three different brands. In fact, this is to test three accelerometers, one from each smartphone.

 Piyasak Thiandee, Boonsap Witchayangkoon, Sayan Sirimontree, Ponlathep Lertworawanich

This research concerns vibration detection while driving and assessment of the relationships between International Roughness Index (IRI) and acceleration data from vibration obtained from three smartphones. Table 1 gives detail of the three models of mobile phones used in this research study. With 100 Hz capability, RSS App gets 100 acceleration data for every one second from vibration, and the average value is recorded.

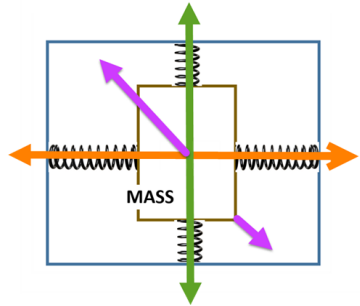

**Figure 1**: Model of accelerometer within the smartphone.

**Table 1**: Mobile phones used in this study

| Mobile Model | HONOR 10 | NOKIA 7 PLUS | SAMSUNG GALAXY A8 (2018) |
|---|---|---|---|
| Operating System | ANDROID 8.1 | ANDROID 8.0 | ANDROID 7.1.1 |
| Processing Unit | KIRIN 970 | Qualcomm Snapdragon 660 | Exynos 7885 |
| Accelerometer | ACC_LSM6DS3-C | BMI160 | LSM6DSL |

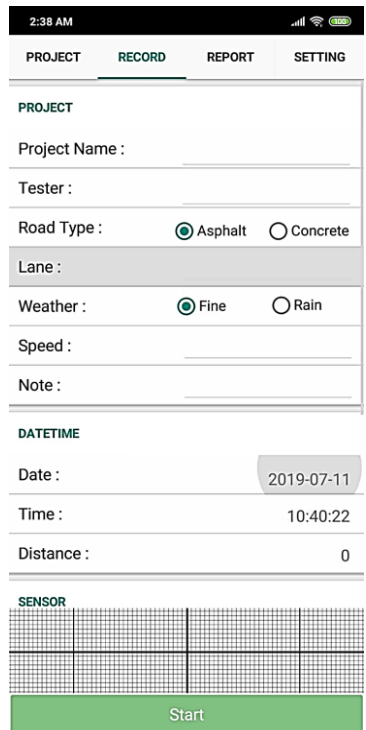

**Figure 2**: RSS App for Android® Smartphones

## 3.2 STANDSTILL MODE SETUP

Before the field-testing, all the three mobile phones have been brought to a standstill and tested for vibration. With the recorded movement of the mobile phones, a simple vibration is observed through the use of Root Mean Square ($RMS = \sqrt{x^2 + y^2 + z^2}$). The obtained result that came up was approximately 9.4~10.0 (range of Gravity value, 9.81 m/s$^2$).

## 3.3  SMARTPHONE SHAKING SETUP

Manual shaking is done for all three smartphones to check validity of sensors of acceleration and stabilization of RSS App.   The output is then calculated with the RMS method.   The obtained result is in the range of 4-6.5m/s$^2$, considered valid for further experiment.

## 3.4  INSTALLATION SETUP OF SMARTPHONES ON THE RUN VEHICLE

Installation of the mobile phones is by adhering all three smartphones to the side-window of the pickup truck.   The installation is make sure that smartphones have free movement for accelerometer to detect vibration.

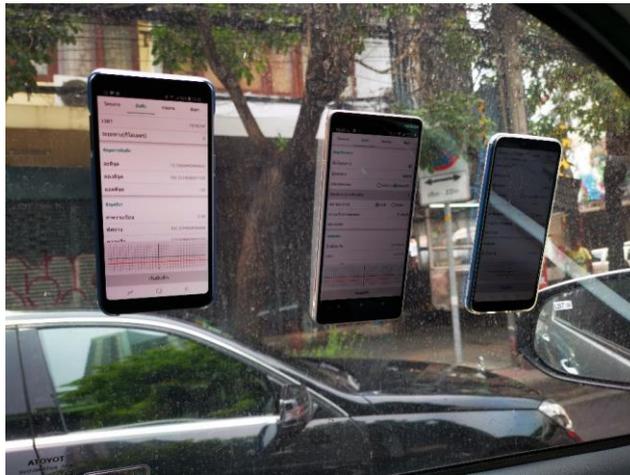

**Figure 3**: Installation of the mobile phones adhered to the window of the pickup truck.

## 3.5  FIELD SURVEY

For field survey, the researchers go out with a pickup truck used for pavement roughness survey with the mobile phones adhering to the car window. The three mobile phones and computer in the car have been connected to Wi-Fi network for time synchronization.

### 3.5.1  TEST ROUTES

During late April 2019, this pre-surveyed study experiments on five different asphaltic concrete routes in the central road network of Thailand, with a distance 15km each.   On May 15, 2019, the sixth and seventh field surveys are conducted on asphaltic concrete road for 9 kilometers long each, on the Si Rat Expressway of Thailand's expressway system.

### 3.5.2  RUN VEHICLE

This test uses a car built according to the standard ASTM E1082-90, see Figure 3.   This car has a laser tool to record road roughness according to the IRI and the time is stamped at all record.   The recorder has been pre-set to record the road roughness every five meters distance.   The car running speed is set at 15m/s.

 Piyasak Thiandee, Boonsap Witchayangkoon, Sayan Sirimontree, Ponlathep Lertworawanich

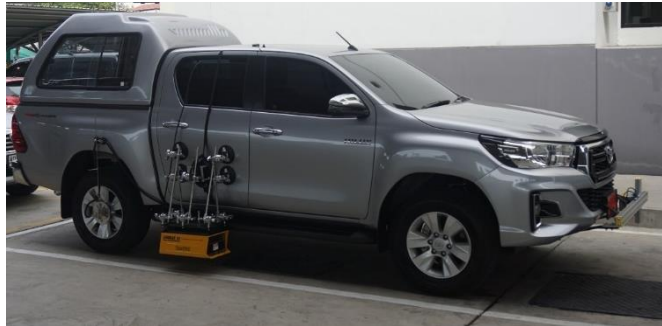

**Figure 4:** the pickup truck used for measurement of IRI in this research.

## 4. RESULT AND DISCUSSION

Before each field experiment, the tests have been conducted to make sure that all smartphones are in good conditions.   The tests include smartphone standstill test (*see* topic 3.2), and smartphone shaking setup (*see* topic 3.3).   After starting the RSS App, the RSS App system is re-set to zero for all values of acceleration.   This paper will pay attention to the result from the sixth and seventh field surveys.

### 4.1 FIELD RESULT

The car laser tool records the road roughness (IRI value) every five meters distance, with vehicle speed 15 m/s.   In one second, the system obtained three IRI values and the average IRI was recorded. The average recorded value of acceleration data from vibration is extracted from smartphones.   The IRI data and acceleration data is linked to the same second.   Each field survey gives more than 600 data pair.

### 4.2 ROOT MEAN SQUARE

The basic calculation is to use the root mean square method, that $RMS = \sqrt{x^2 + y^2 + z^2}$, where $x, y, z$ are acceleration data from vibration in x, y, and z directions, respectively. The RMS result is used to find a relationship with IRI.

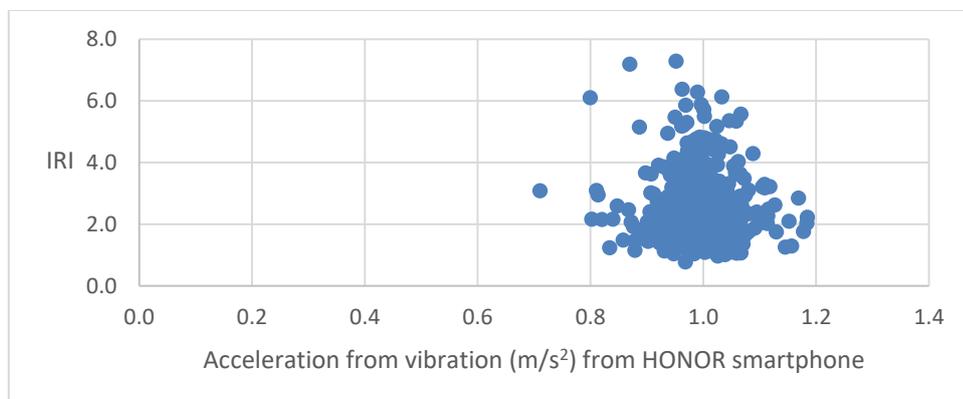

**Figure 5**: Relationship between acceleration from vibration (m/s$^2$) and IRI values from HONOR 10 smartphone, driving speed 15m/s.

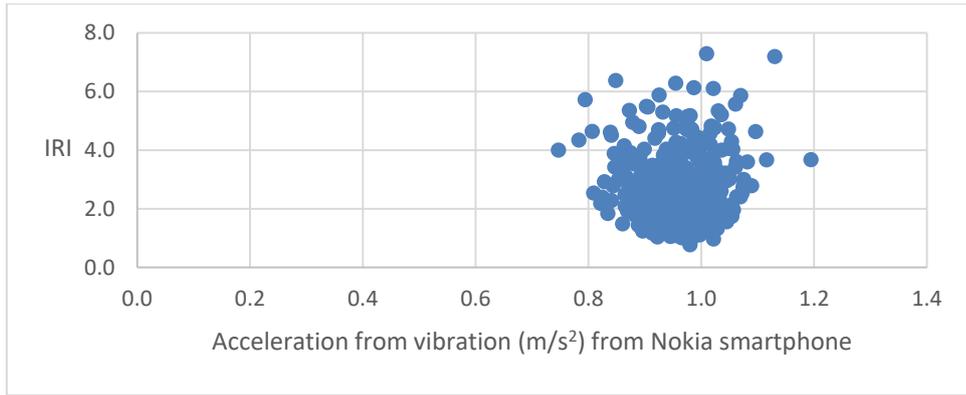

**Figure 6**: Relationship between acceleration from vibration (m/s$^2$) and IRI values from NOKIA 7 PLUS smartphone, driving speed 15m/s.

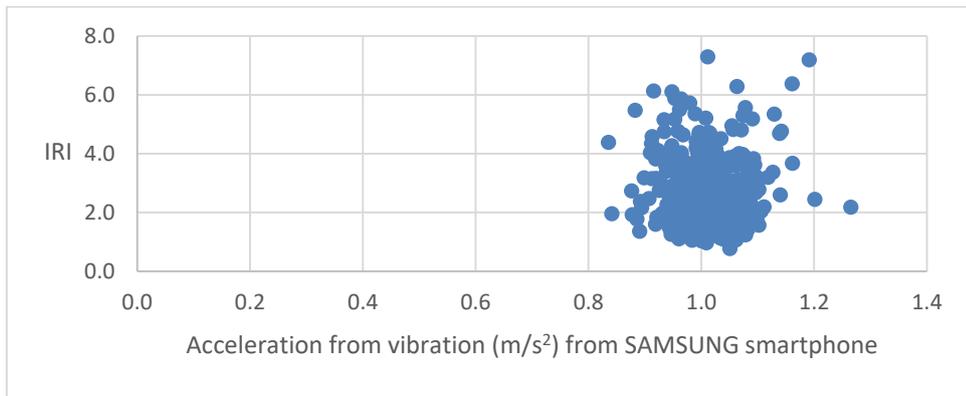

**Figure 7**: Relationship between acceleration from vibration (m/s$^2$) and IRI values from SAMSUNG GALAXY A8 (2018) smartphone, with speed 15m/s.

From Figures 5, 6, and 7, with varied IRI values, it can see that the accelerations from vibration of all tested smartphones are in the range 0.8-1.2 m/s$^2$. This seems to be the limit of accelerometers of the smartphones, resulted in very little correlations between acceleration from vibration (m/s$^2$) and IRI values.

### 4.3 MACHINE LEARNING

Machine learning (ML), the computer-based scientific technique, can help to calculate to covert the smartphones' acceleration data from vibration to IRI. With supervised ML algorithms and statistical concept, the patterns in the 2km long acceleration data is used as a training data. After the training, the predictions for the rest of 7km long acceleration data is possible. This result from ML can give the relationship between acceleration data from vibration and IRI. The study applies the TAN-SIGMOID Transfer Function $y$,

$$y = \frac{2}{1 + e^{-2a}} - 1 \tag{1},$$

$$a = b + \sum_{i=1}^{n} P_i W_i \tag{2},$$

where $W$ is weight, $b$ is a constant, and $P$ is acceleration value from the smartphones, for data time $i$ to time $n$. The process is given in Figure 4.

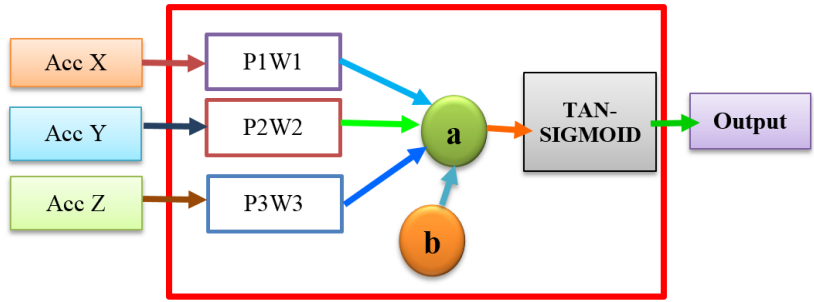

**Figure 8**: ML process based on Tan Sigmoid transfer function (Artificial Neural Networks).

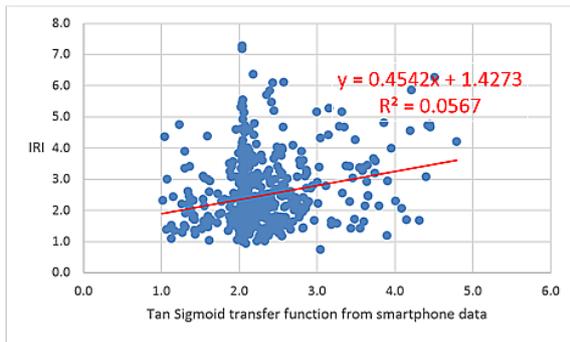 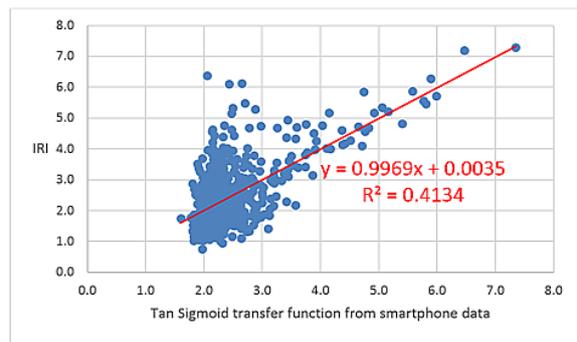

(a) 2 km training                           (b) 9 km training

**Figure 9**: TAN-SIGMOID Transfer Function from HONOR smartphone data.

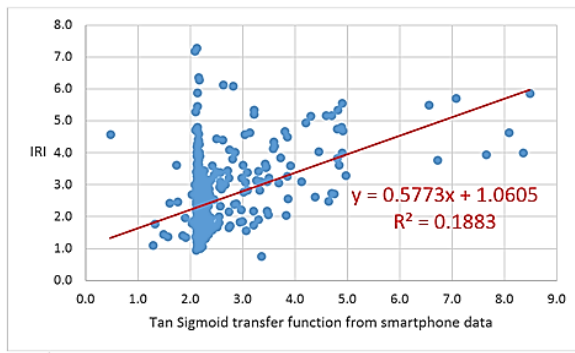 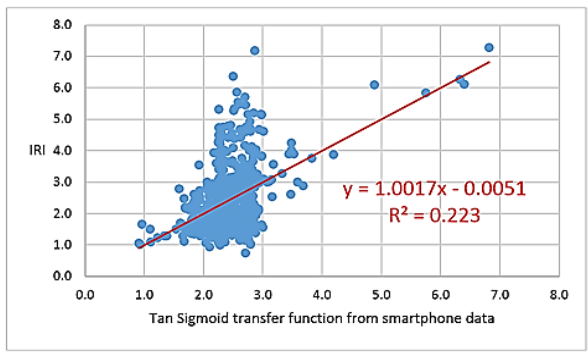

(a) 2 km training                           (b) 9 km training

**Figure 10**: TAN-SIGMOID Transfer Function from NOKIA smartphone data.

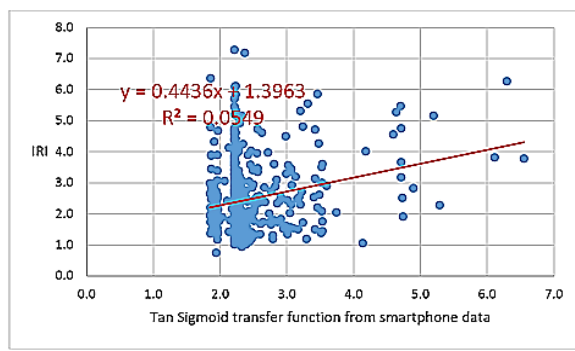 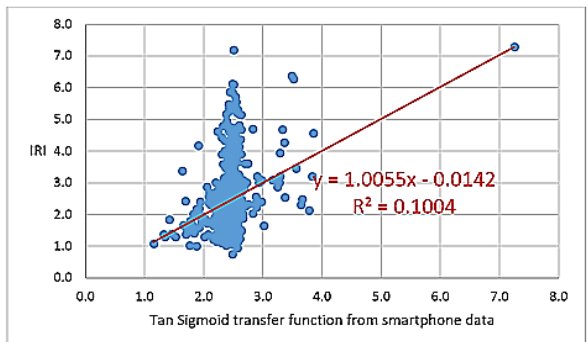

(a) 2 km training                           (b) 9 km training

**Figure 11**: TAN-SIGMOID Transfer Function from SAMSUNG smartphone data.

ML method via TAN-SIGMOID Transfer Function gives better result for all cases of trainings,

see Figures 9-11. For the 2 km training, Figures 9(a), 10(a), and 11(a), the analysis results for 7km are a lot better compared to RMS. In Figures 9(b), 10(b), and 11(b), the 9 km training, the analysis for the same 9km data, the ML does not give perfect results due to low relationships between IRI and acceleration data from vibration extracted from smartphones.

When consider these low, the acceleration data from vibration seem to stay in the range 0.8-1.2 $m/s^2$, no matter the IRI values are, even though the setup shakings give acceleration data result 4-6.5 $m/s^2$. The reason may due to the installation in the vertical that force the smartphones to move in only one direction, while the shaking test is done in three axes.

## 4.4 CORRELATIONS BETWEEN EXPERIMENT OUTCOME AND IRI

The correlations between IRI values and acceleration data from vibration from smartphones are shown in the Table 2. It can see that each smartphone give different result while SAMSUNG smartphone gives the worst correlation.

**Table 2**: Correlations between IRI and acceleration data from vibration.

| Calculation method | HONOR | NOKIA | SAMSUNG |
|---|---|---|---|
| RMS | 0.001 | 0.000 | 0.000 |
| ML from 2km training | 0.057 | 0.188 | 0.055 |
| ML from 9km training | 0.413 | 0.223 | 0.100 |

## 5. CONCLUSION

The study present the experiment on using three different smartphones to collect acceleration data from vibration to detect the road roughness through the RSS App in the Android operating system. The RMS and ML methods are used in this study. There is different ability of each smartphone to detect the vibration, thus different detected values are obtained. The results are compared to the IRI observation. ML method gives better result compared to RMS. This study found no relationship on between IRI and acceleration data from vibration collected smartphones.

## 6. DATA AVAILABILITY STATEMENT

The used or generated data from this study are available upon request to the corresponding author.

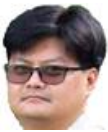

**Piyasak Thiandee** is a Master's student at Department of Civil Engineering, Thammasat School of Engineering, Thammasat University. He got his Bachelor's degree in Civil Engineering from Ramkhamhaeng University, Thailand. His research is related to Transportation Management.

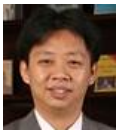

**Dr. Boonsap Witchayangkoon** is an Associate Professor in Department of Civil Engineering, Thammasat School of Engineering, Thammasat University, Thailand. He received his B.Eng. from King Mongkut's University of Technology Thonburi with Honors. He earned his PhD from University of Maine, USA in Spatial Information Science & Engineering. Dr. Witchayangkoon current interests involve applications of emerging technologies to engineering.

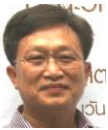

**Dr. Sayan Sirimontree** earned his bachelor degree from Khonkaen University Thailand, master degree in Structural Engineering from Chulalongkorn University Thailand and PhD in Structural Engineering from Khonkaen University Thailand. He is an Associate Professor at Thammasat University Thailand. He is interested in durability of concrete, repair and strengthening of reinforced and prestressed concrete structures.

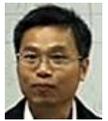

**Dr.Ponlathep Lertworawanich** is Director of the Bureau of Highways Maintenance Management, Department of Highways, Ministry of Transportation, Royal Thai Government, Thailand. He earned his Bachelor of Engineering degree from Chulalongkorn University, Thailand. He obtained his PhD from The Pennsylvania Transportation Institute, The Pennsylvania State University, USA. His researches include Traffic Problem, Traffic Management and Statistics.